%% file: main.tex
\documentclass[letterpaper, 10 pt, conference]{ieeeconf}  % Comment this line out if you need a4paper

\IEEEoverridecommandlockouts 

\usepackage{graphicx}
\usepackage{glossaries}
\usepackage{amsfonts}
\usepackage{caption}
\usepackage{subcaption}
\usepackage{tabularx}

\usepackage{amsmath}

\DeclareMathOperator*{\argmin}{arg\,min}
\newcommand{\mb}[1]{\mathbf{#1}}

\begin{document}

\title{\LARGE \bf Magnetic Ball Chain Robots for Endoluminal Interventions}

\author{Giovanni~Pittiglio, Margherita~Mencattelli and Pierre~E.~Dupont 
\thanks{The authors are with the Department of Cardiovascular Surgery, Boston Children’s Hospital, Harvard Medical School, Boston, MA 02115, USA. Email:
        {\tt\small \{giovanni.pittiglio, margherita.mencattelli, pierre.dupont\}@childrens.harvard.edu} \newline This work was supported by the National Institutes of Health under grant R01HL124020.}%
}

\maketitle

\begin{abstract}
This paper introduces a novel class of hyperredundant robots comprised of chains of permanently magnetized spheres enclosed in a cylindrical polymer skin. With their shape controlled using an externally-applied magnetic field, the spherical joints of these robots enable them to bend to very small radii of curvature. These robots can be used as steerable tips for endoluminal instruments. A kinematic model is derived based on minimizing magnetic and elastic potential energy. Simulation is used to demonstrate the enhanced steerability of these robots in comparison to magnetic soft continuum robots designed using either distributed or lumped magnetic material. Experiments are included to validate the model and to demonstrate the steering capability of ball chain robots in bifurcating channels.
\end{abstract}

\begin{keywords}
Medical Robots and Systems, Steerable Catheters/Needles, Flexible Robotics, Magnetic Actuation.
\end{keywords}

\IEEEpeerreviewmaketitle
\section{Introduction}
\input{sections/introduction}

\section{Robot Kinematics}
\label{sec:kinematics}
\input{sections/kinematics}

\section{Comparison with Designs Using Discrete Magnets and Distributed Particles}
\label{sec:comparison}
\input{sections/comparison}

\section{Experiments}
\label{sec:validation}
\input{sections/validation}

\section{Conclusions}
\input{sections/conclusions}

%\section*{Acknowledgment}

\bibliographystyle{IEEEtran}
\bibliography{bibliography}											
\end{document}

%% file: sections/introduction.tex
%\IEEEPARstart{T}{he} 
Continuum and discrete-jointed robots are well suited to minimally invasive medical interventions \cite{Dupont2022}. Their shape and flexibility enables them to be inserted into the body either percutaneously or through a natural orifice and from there to be navigated through endoluminal passages to targets deep within the body. Applications include endovascular and intracardiac catheters to treat cardiovascular diseases, bronchoscopic robots to diagnose lung cancer and colonoscopes to treat bowel disease. With total lengths up to about one meter, these robots are typically constructed of a long flexible passive proximal section and a short distal steerable section. 

\begin{figure}[!ht]
    \centering
    \includegraphics[width=\columnwidth]{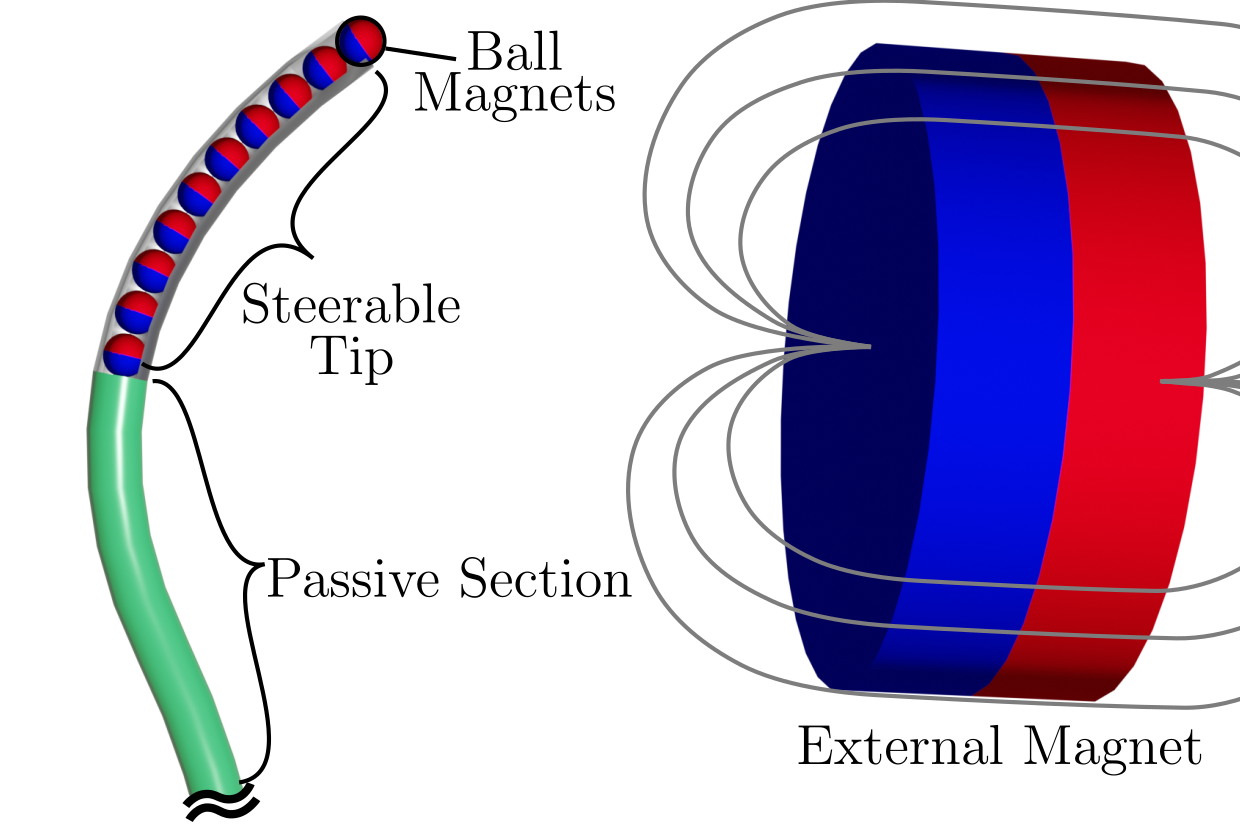}
    \caption{Magnetic ball chain robot as the steerable tip of an interventional system.}
    \label{fig:robot_description}
\end{figure}
For controlling the shape of the steerable tip, magnetic actuation offers several advantages \cite{Abbott2020a}. In this approach, an external magnetic field is applied which generates torques on permanently magnetized material embedded in the steerable section. Since magnetic fields are transmitted through the body, no mechanical or electrical power needs to be transmitted through the proximal length of the robot. This results in robot designs that are smaller in diameter, less complex and less costly \cite{daVeiga2020a}. 

While substantial research effort has been invested in how to generate the external magnetic field, e.g., with combinations of electromagnetic coils \cite{Salmanipour2021, Fischer2022, Heunis2020, Yang2021} or robot-mounted permanent magnets \cite{Pittiglio2019, Barducci2019, Pittiglio2020a, Pittiglio2022b}, the systematic design of the steerable tip is a topic of current and growing interest \cite{Wang2021, Pittiglio2022}. The earliest magnetic steerable tip designs were comprised of a single permanent magnet attached to the distal end of a compliant composite or polymer rod \cite{Chun2007}. Variations of this design incorporate one or several additional permanent magnets embedded along the length of the rod \cite{Edelmann2017, Jeon2018}. More recently, designs have replaced discrete magnets with magnetic particles mixed into a soft polymer \cite{Kim2019, Kim2022, Pittiglio2022}. 

In designing a magnetically-actuated steerable tip, the tradeoff is that adding magnetic material to the rod increases the magnetic torque generated on it by the external magnetic field, but it also increases the bending stiffness of the rod. Discrete magnets embedded in the rod make it effectively rigid over the magnet's length. Similarly, when magnetic particles are mixed into a soft polymer, the bending stiffness of the resulting composite material increases in proportion to the concentration of the magnetic particles \cite{Wu2019}. 

To address this tradeoff, this paper introduces an alternative design concept that attempts to combine the best features of the lumped and distributed particle designs. In this alternative approach, the magnetic material is lumped, but takes the form of a kinematic chain (Fig. \ref{fig:robot_description}). This technique enables the steerable section to incorporate a large volume of magnetic material while allowing for flexibility along its length.

The kinematic chain is comprised of a set of permanently magnetized NdFeB spheres which self-assemble such that their dipole axes are aligned. The spheres are covered by a thin cylindrical polymer skin to seal out surrounding fluids. The chain acts as series of spherical joints with adjacent spheres capable of rolling, sliding and spinning with respect to each other. Each sphere experiences torques from its neighbors which attempt to keep their dipole axes aligned. This produces the equivalent of a bending stiffness for the mechanism. An external magnetic field produces a torque on each sphere which is directed to align the dipole axis of the sphere with the field. The combination of these effects produce steerable tips capable of high bending curvatures.

The remainder of the paper is arranged as follows. The next section proposes and derives a kinematic model for magnetic ball chain robots. In Section \ref{sec:comparison}, simulation is used to compare ball chains with prior designs using discrete magnets and distributed particles. Experimental validation of the model is provided in Section \ref{sec:validation}. Conclusions appear in the final section of the paper.

%% file: sections/kinematics.tex
\begin{figure}[t]
    \centering
    \includegraphics[width=\columnwidth]{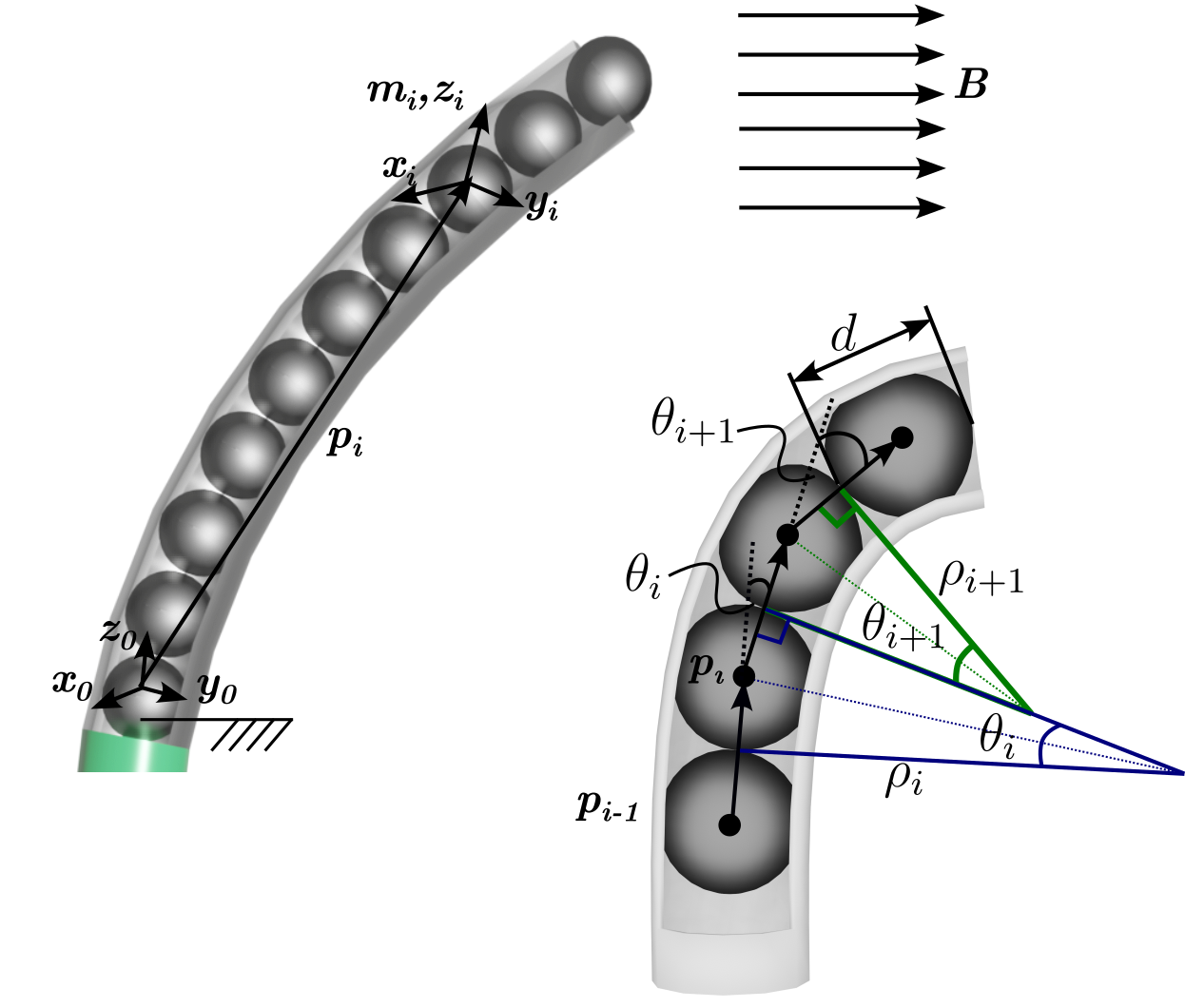}
    \caption{Schematic of ball chain coordinate frames.}
    \label{fig:detailed_description}
\end{figure}

A ball chain robot is shown schematically in Fig. \ref{fig:detailed_description}. It is composed of spherical permanent magnets that naturally align with and attract each other into a linear structure when no external magnetic field is applied. When an external field is applied the spheres experience a torque acting to align their magnetic dipole ($\mb{m}_i$) with the magnetic field ($\mb{B}$). The shape of the steerable tip results from a static equilibrium balance of the torques between the spheres, those produced by the external field, the torque needed to bend the elastic sleeve covering the spheres and by any other externally applied forces or torques. This shape corresponds to the minimum energy configuration of the system. To solve for this shape, we derive an expression for total system energy and use numerical techniques to minimize energy over the space of configuration variables. 

The assumptions made are as follows. We assume that the external magnet produces a uniform magnetic field on the robot. The only other external loading we consider is that due to gravity. Given these assumptions, the spheres always remain in chain form. We also assume that there is no friction between the spheres and between the spheres and the elastic sleeve. While the spheres exactly match the magnetic dipole model \cite{Petruska2013}, we also assume that the external magnet can be approximated as a dipole. The deformation of the elastic sleeve is assumed to be piecewise constant curvature and any twisting is neglected. Finally, to simplify model presentation, we consider only the steerable tip and model it as having its base fixed in a world coordinate frame. 

We define a body frame in each sphere in which the $z$ axis is aligned with its magnetic dipole. In a world frame, $\{x_0,y_0,z_0\}$, the configuration of each sphere can be described by the position, $\mb{p}_i \in \mathbb{R}^3$ and orientation, $R_i \in SO(3)$ of its body frame. A ball chain with $n$ spheres nominally possesses $6n$ degrees of freedom. Since it is assumed that the balls of the chain remain in contact, however, this reduces the independent positional coordinates by $n-1$. Furthermore, magnetic and gravitational potential energy are invariant to rotations of the spheres about their dipole $z$ axes. Finally, for convenience we can locate the proximal sphere so that its body frame coincides with the world frame. Consequently, the total number of independent configuration parameters is $4n-4$.

The magnetic field at a point $\mb{r}\in \mathbb{R}^3$ generated by a dipole located at the origin of magnetic moment $\mb{m}_1=|\mb{m}_1|\hat{\mb{m}}_1\in \mathbb{R}^3$ is given by \cite{Petruska2013}
\begin{equation}
    \label{eq:dipole}
    \mb{B}(\mb{r}, \mb{m}_1) = \frac{|\mb{m}_1| \mu_0}{4 \pi |\mb{r}|^3} \left(3 \hat{\mb{r}} \hat{\mb{r}}^T - I \right) \hat{\mb{m}}_1
\end{equation}
in which, for any vector $\mb{v}\in\mathbb{R}^3$, $\hat{\mb{v}} = \mb{v}/|\mb{v}|$, with $|\cdot|$ the Euclidean norm. 

The magnetic potential energy, $U_d$, of a dipole of moment $\mb{m}_2$ in a magnetic field, $\mb{B}$ is given by
\begin{equation}
    \label{eq:energy}
    U_d = - \mb{m}_2 \cdot \mb{B}
\end{equation}

Using these equations, we can write the potential energy of the ball chain, $U_b$, as the sum of energies from each pair of balls acting on each:
\begin{eqnarray}
U_{b} &=& -\sum_{i = 1}^n \sum_{j = i + 1}^n {} \mb{m}_j \cdot {} \mb{B}(\mb{p}_j - \mb{p}_i, \mb{m}_i )  \\
&=& -\sum_{i = 1}^n \sum_{j = i + 1}^n {} |\mb{m}_j|{} \left(R_j e_3 \right) \cdot \mb{B}(\mb{p}_j -  \mb{p}_i, |\mb{m}_i|{} \left(R_i e_3 \right) ) \nonumber
\end{eqnarray}
Note that the energy is expressed with respect to the ball chain configuration parameters, $(\mb{p}_i,R_i)$ by using the fact that $\mb{m}_i=|\mb{m}_i| \left(R_i e_3 \right)$.

Considering a cylindrical external magnet of magnetic moment $\mb{m}_e$ and configuration $(\mb{p}_e,R_e)$ with diameter $D$ larger than its length, we can position the magnet such that all the balls in the chain satisfy $|\mb{p}_i - \mb{p}_e| > D, \ i = 1, \dots, n$. In this case, we can model the external magnet as a dipole with an error less than 2\% \cite{Petruska2013}. The potential energy associated with the external magnet acting on the ball chain is given by the following sum.

\begin{equation}
U_{e} = -\sum_{i = 1}^n {} \mb{m}_i \cdot {} \mb{B}(\mb{p}_i - \mb{p}_e, \mb{m}_e ) 
\end{equation}

While we anticipate that in many cases, the elastic polymer skin covering the ball chain will be designed to be very flexible and could potentially be neglected in the kinematic modeling, we include its elastic energy due to bending here. We discretize the skin into arcs spanning the contact points between adjacent pairs of balls (Fig. \ref{fig:detailed_description}). The center line of the cylindrical skin section for ball $i$ is assumed to have a constant radius of curvature, $\rho_i$, and to span an angle, $\theta_i$. 

These variables can be expressed in terms of the shape parameters. The angle $\theta_i$ is defined by
\begin{equation}
\tan(\theta_i) = \frac{(\mb{p}_{i + 1}-\mb{p}_i) \times (\mb{p}_{i}-\mb{p}_{i-1})}
{(\mb{p}_{i + 1}-\mb{p}_i) \cdot (\mb{p}_{i}-\mb{p}_{i-1}) }
\label{eq:theta}
\end{equation}
and the arc length is given by
\begin{equation}
\rho_i = (d/2)\cot(\theta_i/2)
\label{eq:rho}
\end{equation}

The elastic strain energy in a beam is given by
\begin{equation}
U = \int_0^L \frac{M^2 ds}{2EI} 
\end{equation}
in which $L$ is the length of the beam, $M$ is the bending moment, $E$ is the elastic modulus and $I$ is the area moment of inertia for the cross section. For a beam experiencing pure bending with radius of curvature, $\rho$, the moment $M$ is constant and given by
\begin{equation}
M=EI/\rho
\end{equation}
For the skin segment covering ball $i$, we can combine the preceding equations and, noting that the length of the segment is given by $\rho_i \theta_i$, obtain an expression for its strain energy, $U_s(i)$,
\begin{equation}
U_s(i) = \frac{E_i I_i \theta_i}{2 \rho_i}
\end{equation}
Here, it is assumed that the elastic modulus, $E_i$, and second moment of area, $I_i$, are piecewise constant.
The total strain energy in the skin due to bending is obtained by summing over the ball segments and noting that this expression is a function of the shape parameters based on equations (\ref{eq:theta}) and (\ref{eq:rho})
\begin{equation}
\label{eq:elastic}
U_s = \frac{1}{2}\sum_{i = 1}^n \frac{E_i I_i \theta_i}{\rho_i}
\end{equation}

Finally, we can define the gravitational potential energy of the ball chain as 
\begin{equation}
U_g = \sum_{i = 1}^n \mu_i \mb{g}^T \mb{p}_i,
\end{equation}
with $\mu_i$ as the mass of the $i$th ball and $\mb{g}$ as the gravitational acceleration vector.

The expression for total potential energy, $U$ is given by the sum of magnetic, elastic and gravitational energies:
\begin{equation}
    \label{eq:equilibrium}
    U = U_b + U_e + U_s +U_g.
\end{equation}
These terms are all functions of the kinematic shape parameters, $(\mb{p}_i, R_i), \ i \in 2, \ldots, n$ and of the input parameters, $(\mb{p}_e, R_e)$. 
Robot shape corresponds to the set of shape parameters, $(\mb{p}_i^*, R_i^*), \ i \in 2, \ldots, n$, that minimize potential energy, $U$, for a given set of input parameters:
\begin{equation}
(\mb{p}_i^*, R_i^*), \ i \in 2, \ldots, n = \argmin_{(\mb{p}_i, R_i), \ i \in 2, \ldots, n} U
\end{equation}
Recall that the shape parameters are constrained since the balls remain in contact and also that shape is independent of the rotation of any sphere about an axis through its center. Consequently, the problem can be solved as a constrained function minimization using routines such as {\it fmincon} from Matlab (Mathworks, Natick, MA).

%% file: sections/comparison.tex
To evaluate its relative steerability, the ball chain design was compared via simulations with two existing alternatives, a single magnet attached to the tip of a flexible polymer rod and a composite rod incorporating magnetic particles. To make the comparison as fair as possible, the outer diameter of all designs was set to be 1mm and the same silicone polymer was used in each design. The properties of all designs are given in Table \ref{table:designscomparison}.

%\begin{table}
%\vspace{0.2cm}
%\centering
%\caption{\label{tab:samples_list} Steerable tip design properties}
%\label{table:properties}
%\begin{tabular}{lccccc}  
%\toprule
%\multicolumn{4}{r}{Parameters} \\
%\cmidrule(r){2-6}
%Tube & Transmission & Steerable Distal & ID  & OD  & Tendon \\
%& Length & Length & & & Radius\\
%& (cm) & (mm) & (mm) & (mm) & (mm)\\ 
%\midrule
%Outer & 91.5 &  123 & 7 & 8 & 3.9\\
%Inner & 121.5 & 123 & 6 & 7 & 3.4\\
%\bottomrule
%\end{tabular}
%\end{table}

\subsection{Design Descriptions}
The ball chain robot was comprised of 0.9mm diameter N52 magnetic spheres with a magnetic dipole intensity $|\mb{m}_b| = B_r V_b/\mu_0$ in which $V_b$ is the volume of the sphere. The cylindrical silicone skin has a thickness of 0.05 mm so that the overall diameter is 1mm. 

For the design using a single magnet, a 1mm diameter rod of silicone incorporated a N52 magnet at its tip. The magnet is modeled as cylindrical with a length of 1mm and a diameter of 0.9mm. It is encased in 0.05mm thick silicone to attach it to the rod. With a remanence $B_r$ = 1.48 T, its magnetic dipole intensity is computed as $|\mb{m}_m| = B_r \ V_m/ \mu_0$, with $V_m$ as the volume of the magnet and $\mu_0$ air magnetic permeability. The magnetic dipole direction was assumed aligned with its cylindrical axis. 

For the design using magnetic particles, the work  of da Veiga \emph{et al.} \cite{DaVeiga2021} was used as a source for the experimental properties of a 1mm diameter rod in which silicone is doped with magnetic particles using a weight ratio of 150\%. The magnetic dipole direction is taken to be aligned with the axis of the rod. 

\subsection{Computing Robot Shape}

The shape of the ball chain robot was computed using the method described in the previous section. The shapes of the other two designs were also computed using an energy minimization approach. Specifically, the rod centerlines were discretized into $m$ segments by defining a set of equally-spaced points along their length, $\mb{p}_i, \ i \in 1, \ldots, m$. An additional point can be added, as needed, at the center of the discrete tip magnet. Equations (\ref{eq:theta})-(\ref{eq:elastic}) can then be used to compute total strain energy due to bending in the polymer rods. 

For the robot with a discrete tip magnet, equations (\ref{eq:dipole}) and (\ref{eq:energy}) are adapted to model the magnetic potential energy. In the design with distributed magnetic particles, magnetic potential energy was computed as the sum over all the discrete elements with the magnetic moment of each element directed along the tangent to the center line. Following the approach used in \cite{Wang2021, Lum2016}, magnetic potential energy associated with bending of the rod incorporating magnetic particles was not considered. To clarify the comparison, gravity was not considered, as in \cite{Wang2021}.

\begin{table}
\vspace{0.2cm}
\centering
\caption{\label{tab:designscomparison} Designs comparison}
\label{table:designscomparison}
\begin{tabularx}{\columnwidth}{cccccc}  
\hline \hline
%\multicolumn{4}{r}{Parameters} \\
%\cmidrule(r){2-6}%
 & Ball chain & Tip magnet & \shortstack{Distributed \\ particles} \\
\hline
Polymer & Ecoflex 00-30 & Ecoflex 00-30 &  \shortstack{Ecoflex 00-30 + \\ 150 \%wt NeFeB}\\ \hline
Elastic Modulus & 42.7 kPa & 42.7 kPa &  128.2 kPa\\
 & (skin) & (rod) & (rod) \\ \hline
Remanence & 1.48 T & 1.48 T & 0.59 T \\ 
 & (balls) & (tip cylinder) & (rod) \\ \hline 
Magnetic Dipole & 0.45 mA m$^2$ & 0.67 mA m$^2$ & 0.37 Am \\
& (each sphere) & (tip magnet) & (length unit) \\ 
 \hline \hline
%\bottomrule
\end{tabularx}
\end{table}

% \begin{figure}
%     \centering
%     \includegraphics[width = \columnwidth, trim={2cm 2.5cm 2cm 2cm}, clip]{figures/workspace.png}
%     \caption{Workspace (WS) comparison and sampled configurations (Config.) for ball chain robot (Balls), distributed particles (Particles) and lumped tip magnet (Tip Magnet) over varying insertion length (1 - 20 mm), applied field strength (4 - 40 mT) and field direction (0 - 180$^o$) with respect to $x$ axis.}
%     \label{fig:comparison}
% \end{figure}

\subsection{Workspace Comparison}
\begin{figure}[t]
        \includegraphics[width = \columnwidth]{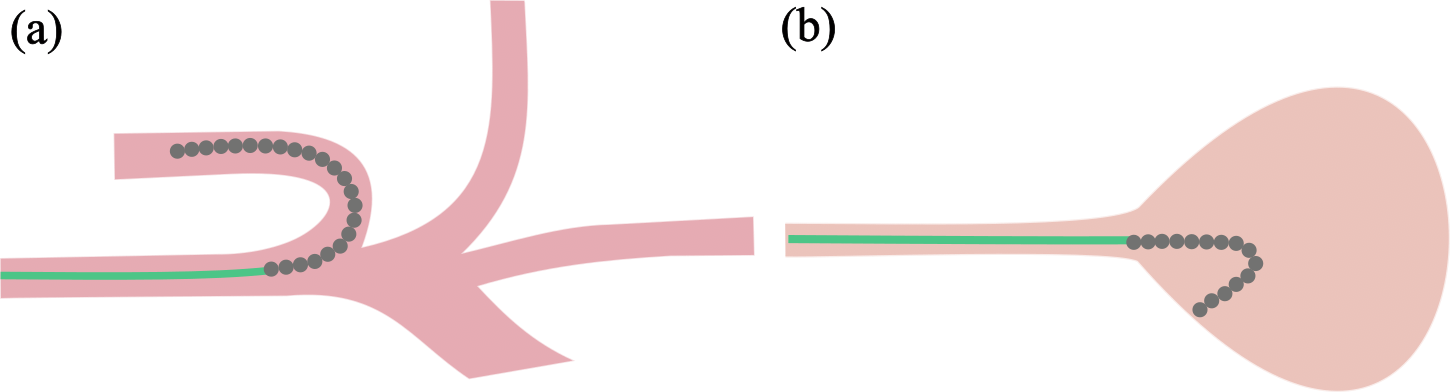}
        \caption{Endoluminal navigation challenges. (a) Steering around a sharp corner in a branching network. (b) Reaching back to access the opening into a body cavity.}
        \label{fig:WS_motivation}
\end{figure}

When comparing designs for steerable robot tips, various metrics can be defined based on task specifications. Of particular interest in endoluminal interventions is the ability to turn tight corners in branching tubular networks, e.g., the vasculature and lungs (Fig. \ref{fig:WS_motivation}(a)). Similarly, when a robot enters a body cavity, e.g., the bladder from the urethra, it is desirable for the robot to be able to reach back to access the opening of the cavity that it entered through, e.g., to treat a lesion at that location (Fig. \ref{fig:WS_motivation}(b)).  

\begin{figure}[htb!]
    \centering
    % \begin{subfigure}[h]{\columnwidth}
        \includegraphics[width = \columnwidth]{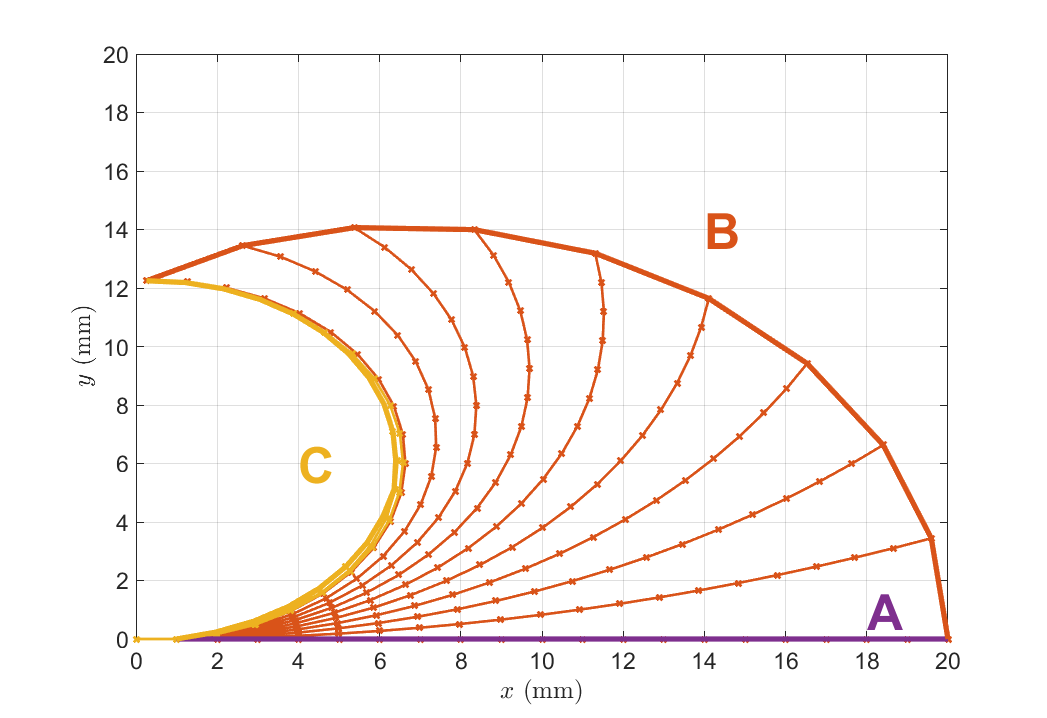}
        \caption{Workspace boundary definition illustrated for the tip magnet design. The planar workspace is defined by 3 boundaries. The first, A, corresponds to the straight configuration as robot length varies from zero to its maximum length. The second, B, corresponds to the external magnetic field rotating by 180 degrees from the positive $x$ to the negative $x$ direction with the robot at its maximum length. The third boundary, C, is generated when the magnetic field is oriented along the negative $x$ axis and the robot length is varied between zero and its maximum length. It is boundary (C) that illustrates the minimum radius of curvature that a design can produce. The steerable tip is able to access all points within the volume obtained when the area between A, B and C is rotated about the $x$ axis.}
        \label{fig:boundary}
        \end{figure}
    \begin{figure}[htb!]
        \includegraphics[width = \columnwidth, trim = {2.5cm, 1cm, 2.5cm, 1.25cm}, clip]{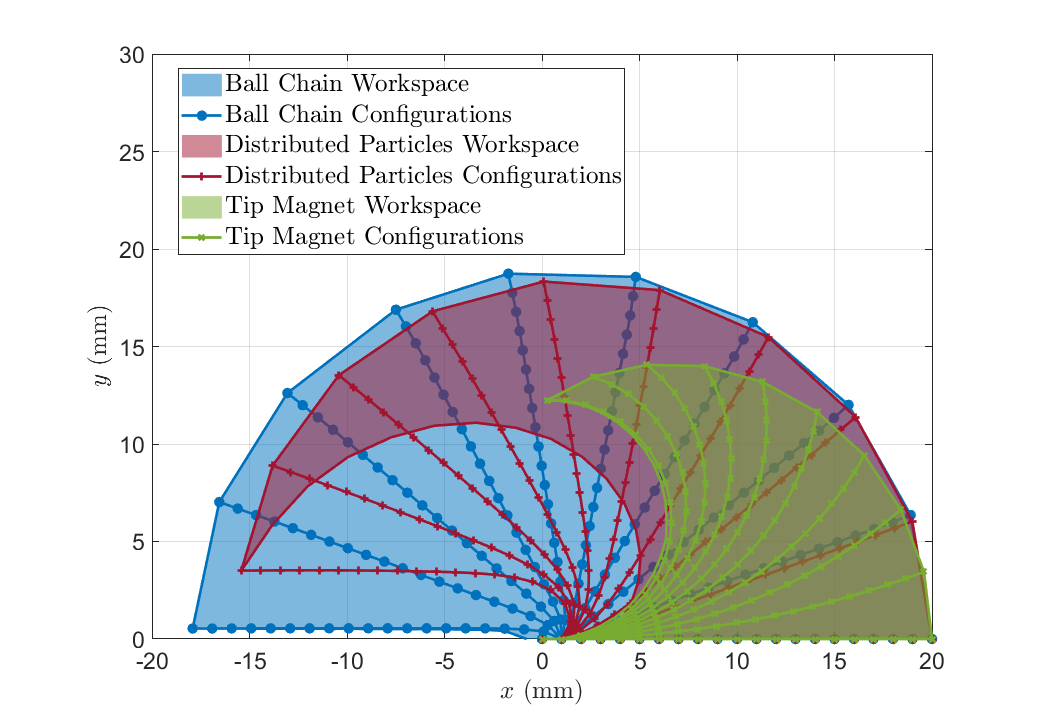}
        \caption{Workspace comparison of ball chain, distributed particle and tip magnet robot designs.}
        \label{fig:comparison}
        %\caption{Robot workspace definition and comparison.}
\end{figure}

These tasks motivate a definition of workspace that not only accounts for the region in front of the robot, but also measures how tight a corner the robot can turn. We define such a workspace as shown in Fig. \ref{fig:boundary} by first solving for the planar set of tip positions that can be reached using the maximum strength magnetic field while varying the field direction over an angle of 0 to 180 degrees with respect to the axis of the robot and while also varying the length of the robot from zero to the full length of the steerable tip. This defines three boundary curves as shown in Fig. \ref{fig:boundary}. This swept area can be rotated about the axis of the robot to produce the swept volume of the workspace. While our workspace definition is different, it was inspired by the approach taken in \cite{Wang2021}.

Fig. \ref{fig:comparison} provides a comparison of the three designs using this workspace definition. A 40 mT homogeneous magnetic field was rotated from parallel (0$^\text{o}$) to anti-parallel (180$^\text{o}$) to the $x$ axis. In addition, robot length was varied in 1 mm steps up to 20 mm. The robot length-to-diameter ratio and field homogeneity were selected based on the work by Wang \emph{et al.} \cite{Wang2021}.

The ball chain design has the largest workspace. Measured in the plane, its workspace is 544 mm$^2$ out of a possible 628.3 mm$^2$ with the latter corresponding to a design with a revolute joint at its base. The workspace areas of the other two designs are 326 mm$^2$ and 170 mm$^2$, respectively.

As experimentally shown in the next section, the capability of the ball chain design to produce a tight turning radius equates to easing navigation in sharply bent lumina and to facilitating exploration of complex anatomical structures.

%% file: sections/validation.tex
\begin{figure}[t]
    \centering
    \includegraphics[width = \columnwidth]{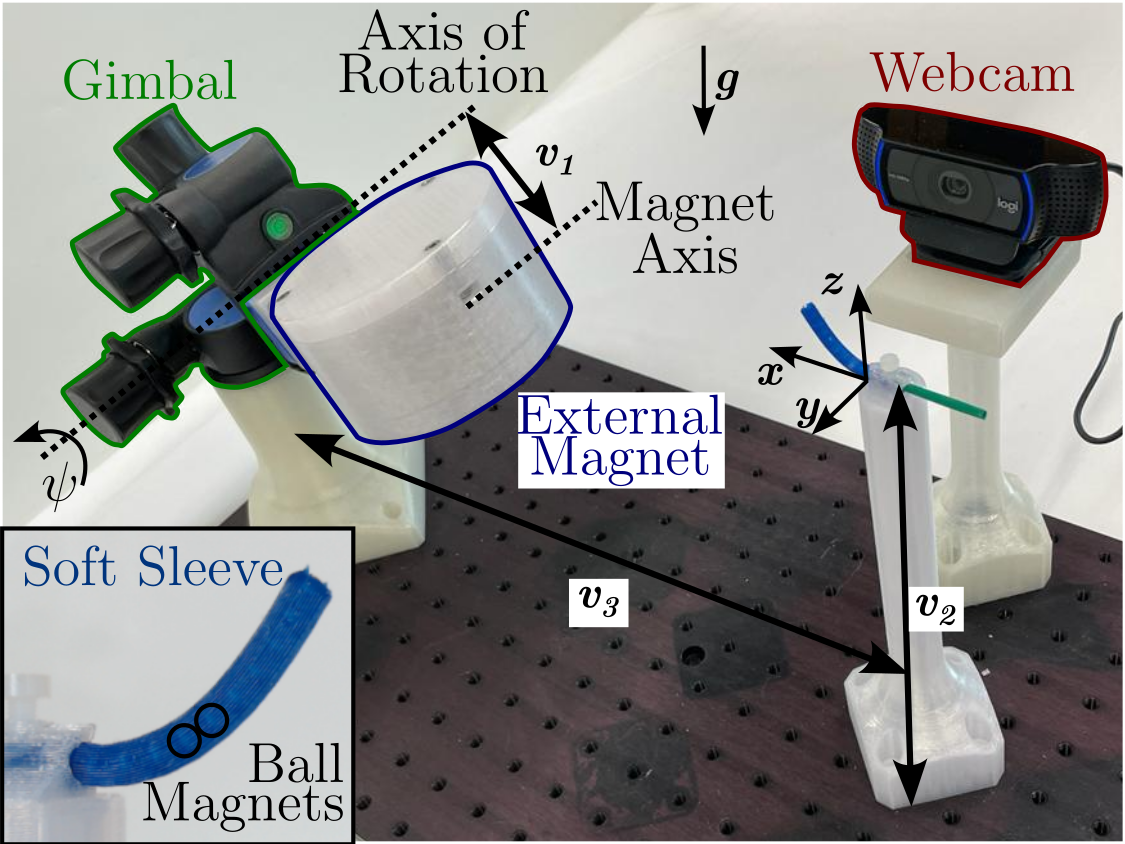}
    \caption{Experimental setup for validation of the proposed ball chain robot model.}
    \label{fig:setup}
\end{figure}

Two experiments were performed to evaluate the ball chain robot design. The first experiment was to validate the proposed kinematic model. The second experiment was designed to evaluate the ability of the robot to turn tight corners in a braching network. Both are described below.

\subsection{Model Validation}

We performed experiments using the set up shown in Fig. \ref{fig:setup}. The robot was composed of ten N42 magnetic spheres of diameter 3.175 mm, mass 0.13 g and remanence 1.32 T. This was the smallest strong spherical permanent magnet available commercially (K\&J Magnetics, USA). Smaller and stronger custom magnets can be fabricated and used in applications which require robots with smaller diameter and higher magnetic content.

An external N52 cylindrical magnet (76.2mm diameter, 38.1mm long, 1.48 T remanence) was mounted on a 3-axis gimbal so that it could be precisely moved with respect to the ball chain robot. Magnet configurations were selected so as to produce a magnetic field in the $x-z$ plane, where gravity was aligned with the negative $z$ axis and the center line of the ball chain was directed along the positive $x$ axis. 

\begin{figure}[t]
    \centering
    \includegraphics[width = \columnwidth]{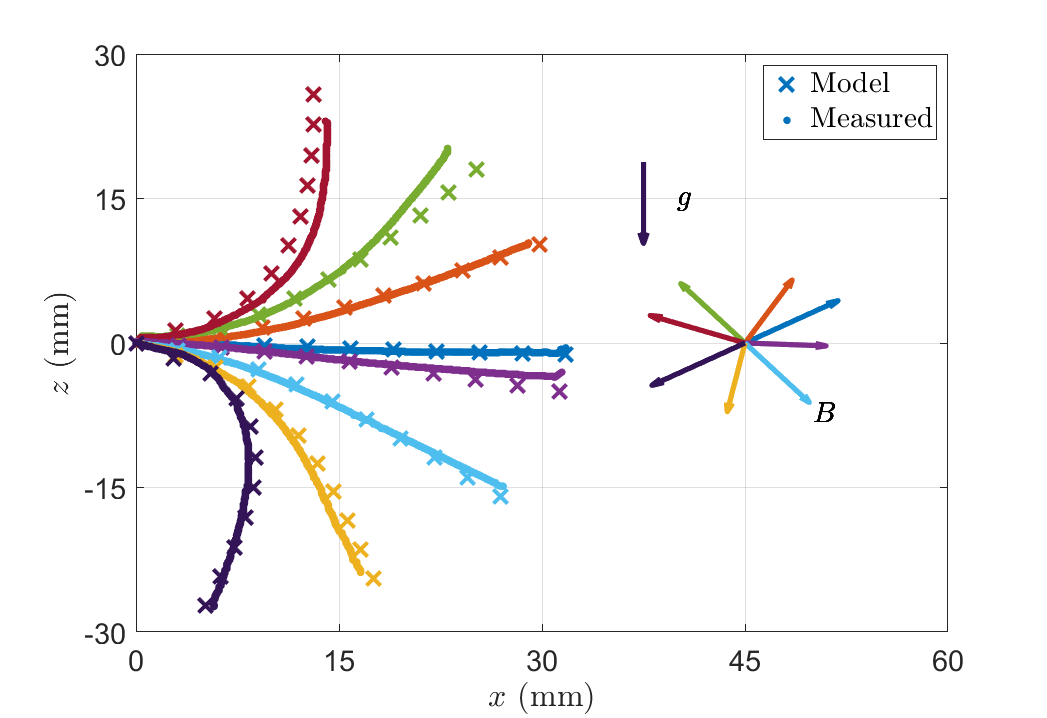}
    \caption{Experimental versus model-predicted shape of the ball chain robot.}
    \label{fig:model_validation}
\end{figure}

A webcam (C920, Logitech, US) was placed by the side of the robot and used to track its center line. The silicone sleeve was molded as combination of Ecoflex 00-30 and blue pigment (Silc Pig, Smooth-on, USA) to produce high visual contrast with the background. The robot's center line was extracted from the blue channel of the images using the \emph{active contour} and \emph{skeletonizing} operations from the Matlab Image Processing Toolbox. This data is reported as ``measured" in Fig. \ref{fig:model_validation}.

The magnet was first positioned with its magnetic dipole direction along the positive $x$ axis ($\psi = 0$), $\hat{\mb{m}}_e = e_1$ and the angle $\psi$ (see Fig. \ref{fig:setup}) increased in steps: $30^\text{o},\ 60^\text{o},\ 75^\text{o},\ 90^\text{o}$. These steps generate the four downward configurations in Fig. \ref{fig:model_validation}. This is the largest rotation we could achieve with the used gimbal. 

Next, the magnet was rotated so that its dipole direction is along the negative $x$ axis, $\hat{\mb{m}}_e = -e_1$. The angle $\psi$ was then adjusted to angles of $90^\text{o},\ 75^\text{o},\ 60^\text{o},\ 30^\text{o}$ to generate the upward motion in Fig. \ref{fig:model_validation}. An extract from the performed experiment can be found in the Supplementary Video.

The magnetic field generated by the external magnet was computed by applying the dipole model to each pose of the external magnet, as expressed in (\ref{eq:dipole}). The position of the magnet with respect to the $i$th ball magnet $\mb{p}_i$ is $\mb{r}_i = \mb{p}_e - \mb{p}_i$, with $\mb{p}_e = (v_3 - v_1 \sin(\psi)) \cdot e_1 + (v_1 \cos(\psi) - v_2) e_3$. Here, $v_1 = 15$ cm, $v_2 = 20$ cm and $v_3 = 35$ cm. 

For each configuration, the shape was computed as described in Section \ref{sec:kinematics}, and compared with the measured shape. The comparison is reported in Fig. \ref{fig:model_validation}. The mean error over all configurations and all spheres in the ball chains is $0.42\pm 0.47$mm.

\subsection{Channel Navigation}

\begin{figure}[t]
    \centering
    \includegraphics[width=\columnwidth]{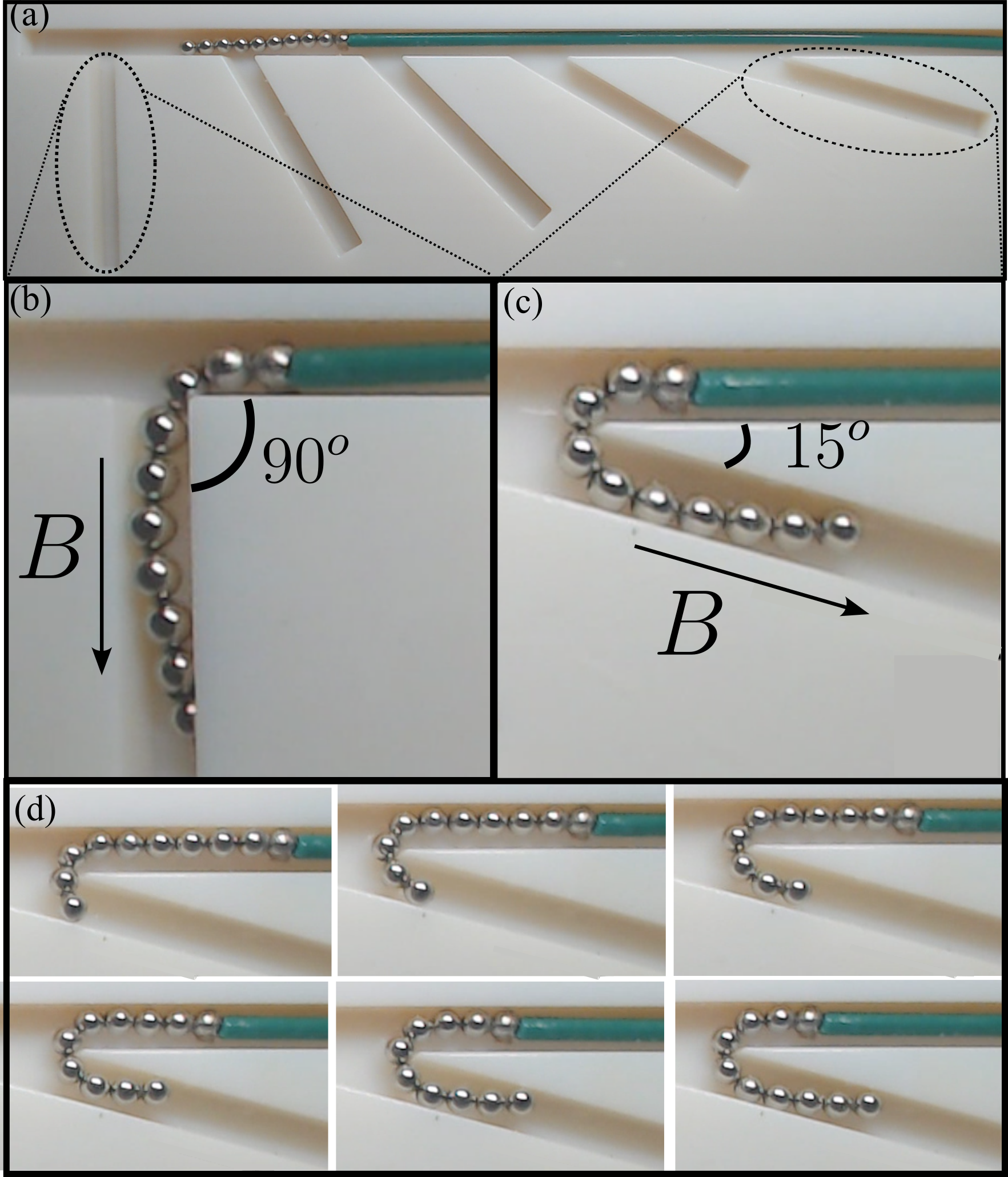}
    \caption{Channel navigation experiment. (a) Set of 5 channels of increasing turning angle. (b) Closeup view of 90$^\text{o}$ turning angle. (c) Closeup view of 165$^\text{o}$ turning angle. (d) Step-wise insertion of case (c).}
    \label{fig:demo}
\end{figure}

To demonstrate the capability for ball chain robots to produce very small radii of curvature, a plate with five different bifurcating channels was 3D printed as shown in Fig. \ref{fig:demo}. The channels are designed with turning angles of 90$^\text{o}$, 120$^\text{o}$, 135$^\text{o}$, 150$^\text{o}$ and 165$^\text{o}$.
A sheet of clear acrylic was placed over the 5mm wide and 5mm deep channels. 

As in the model validation experiments, the robot is comprised of ten 3.175mm diameter N42 magnetic spheres, but in these experiments the steerable tip is mounted to a 2.7mm diameter Pebax tube (Zeus Industrial Products). For visual clarity, the silicone sleeve is omitted in Fig. \ref{fig:demo}, but is included in the Supplemental Video. 

The navigation experiment consisted of manually inserting the semi-rigid tube to reach each bifurcation at which time an external magnet (N52, diameter 76.2 mm, height 38.1 mm) was positioned under the 3D printed plate and manually rotated to align its magnetic field with the direction of each channel. This causes the robot to bend in the desired direction (Fig. \ref{fig:demo}) during continued manual insertion. The applied magnetic field $|\mb{B}|$ was about 40 mT, matching what was used in the simulations of the previous section.  

As shown in Fig. \ref{fig:demo}, the ball chain robot prototype was able to navigate into all of the side channels. The sequence of images in Fig. \ref{fig:demo}(d) demonstrates how the ball chain robot really does behave as a concatenation of spherical joints with most of the robot rotation concentrated in the two to three balls that are passing around the corner. 

%% file: sections/conclusions.tex
The present paper discussed a novel magnetically actuated continuum robot for endoluminal procedures. The novel design is based on a series of ball-chained permanent magnets, which naturally assemble by aligning their magnetic dipole direction. The resulting hyperredundant slender structure is characterized by two main properties: high bending capabilities and high magnetic content. The former is achieved by replacing the continuous bending of a rod with a chain of spherical joints surrounded by a thin flexible skin. In addition, the magnetic mechanism enables the inclusion of of substantially more torque-producing magnetic material in comparison with other designs. 

These properties were demonstrated both by simulating the behaviour of the robot and by validating the model against experimental data. The former underlines the substantial improvement in minimum radius of curvature and in workspace volume attributable to the ball chain design compared to the distributed and lumped magnetic material models already proposed in the literature. 

The proposed model was compared to data extracted from image tracking while applying an external magnetic field by mean of a large permanent magnet. We show sub-millimeter error in the center line tracking. Furthermore, the ability to navigate a branching network with side channels angled between 90$^\text{o}$ and 165$^\text{o}$ was demonstrated. 

The demonstrated steerability suggests that ball chain robots could prove useful in many different types of endoluminal applications (e.g., cardio- and neuro-vascular, pulmonary and gastrointestinal), both for inspection/diagnosis and for treatment. For the delivery of therapeutics, the ball chain robot can be used following the standard approach in which a passive guide wire is first steered into a vascular branch and then a catheter is passed over it. The guide wire is then removed leaving the catheter as a delivery tube.